\documentclass{article}

\usepackage[final]{neurips_2021}

\usepackage[utf8]{inputenc} 
\usepackage[T1]{fontenc}    
\usepackage{hyperref}       
\usepackage{url}            
\usepackage{booktabs}       
\usepackage{amsfonts}       
\usepackage{nicefrac}       
\usepackage{microtype}      
\usepackage{xcolor}

\usepackage{wrapfig}

\usepackage{times}  
\usepackage{helvet} 
\usepackage{courier}  
\usepackage{subcaption, graphicx} 
\usepackage{xparse}
\usepackage{epstopdf}
\usepackage{rotating,multirow}
\usepackage{amsmath,amssymb} 
\usepackage{color}
\usepackage{wrapfig,lipsum}
\usepackage{tabularx}
\usepackage{makecell}

\usepackage{algorithm}
\usepackage{algpseudocode}
\usepackage{amsthm,epsfig}
\usepackage{mathtools}
\usepackage{caption}

\newcommand{\etal}{\textit{et al. }}

\newcommand{\ie}{\textit{i.e., }}

\title{Normality-Calibrated Autoencoder for Unsupervised Anomaly Detection on Data Contamination}

\author{%
  Jongmin Yu$^{1}$, Hyeontaek Oh$^{1}$, Minkyung Kim$^{2}$, and Junsik Kim$^{3}$\\
  $^{1}$Institute for IT Convergence and $^{2}$School of Electrical Engineering\\
  Korea Advanced Institute of Science and Technology (KAIST), Daejeon, 34141, Rep. of Korea\thanks{Email for corresponding author: hyeontaek@kaist.ac.kr} \\
  $^{3}$School of Engineering and Applied Sciences, Harvard University, Cambridge, MA 02138, U.S.A. \\
  \texttt{\{andrew.yu, hyeontaek, mkkim1778\}@kaist.ac.kr, mibastro@gmail.com} \\
}

\begin{document}

\maketitle

\begin{abstract}
In this paper, we propose Normality-Calibrated Autoencoder (NCAE), which can boost anomaly detection performance on the contaminated datasets without any prior information or explicit abnormal samples in the training phase. The NCAE adversarially generates high confident normal samples from a latent space having low entropy and leverages them to predict abnormal samples in a training dataset. NCAE is trained to minimise reconstruction errors in uncontaminated samples and maximise reconstruction errors in contaminated samples. The experimental results demonstrate that our method outperforms shallow, hybrid, and deep methods for unsupervised anomaly detection and achieves comparable performance compared with semi-supervised methods using labelled anomaly samples in the training phase. The source code is publicly available on \url{https://github.com/andreYoo/NCAE_UAD.git}.
\end{abstract}

\section{Introduction}
\label{sec:intro}
Most of anomaly detection (AD) methods \citep{erfani2016,zhai2016,chen2017,ruff2018,deecke2018,ruff2019,golan2018,pang2019,hendrycks2019,hendrycks2019using,zong2018deep} assume that the training dataset only consists of normal samples; however, datasets in real-world are easily \emph{contaminated}, which means that datasets contains both normal and abnormal samples. The contaminated samples significantly degrade the AD performance of models derived based on the assumption.

\begin{figure}
    \centering
     \begin{subfigure}{\textwidth}
    \centering
        \includegraphics[width=0.4\textwidth]{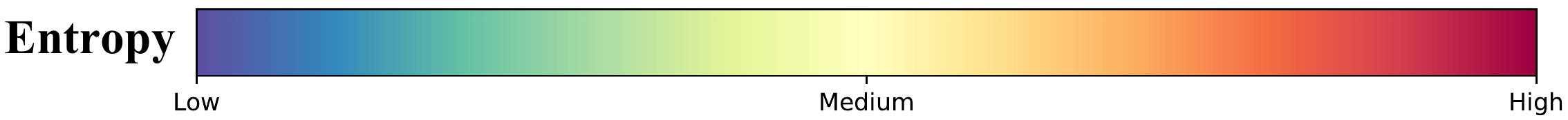}
    \end{subfigure}
    \begin{subfigure}{0.19\textwidth}
        \includegraphics[width=\textwidth]{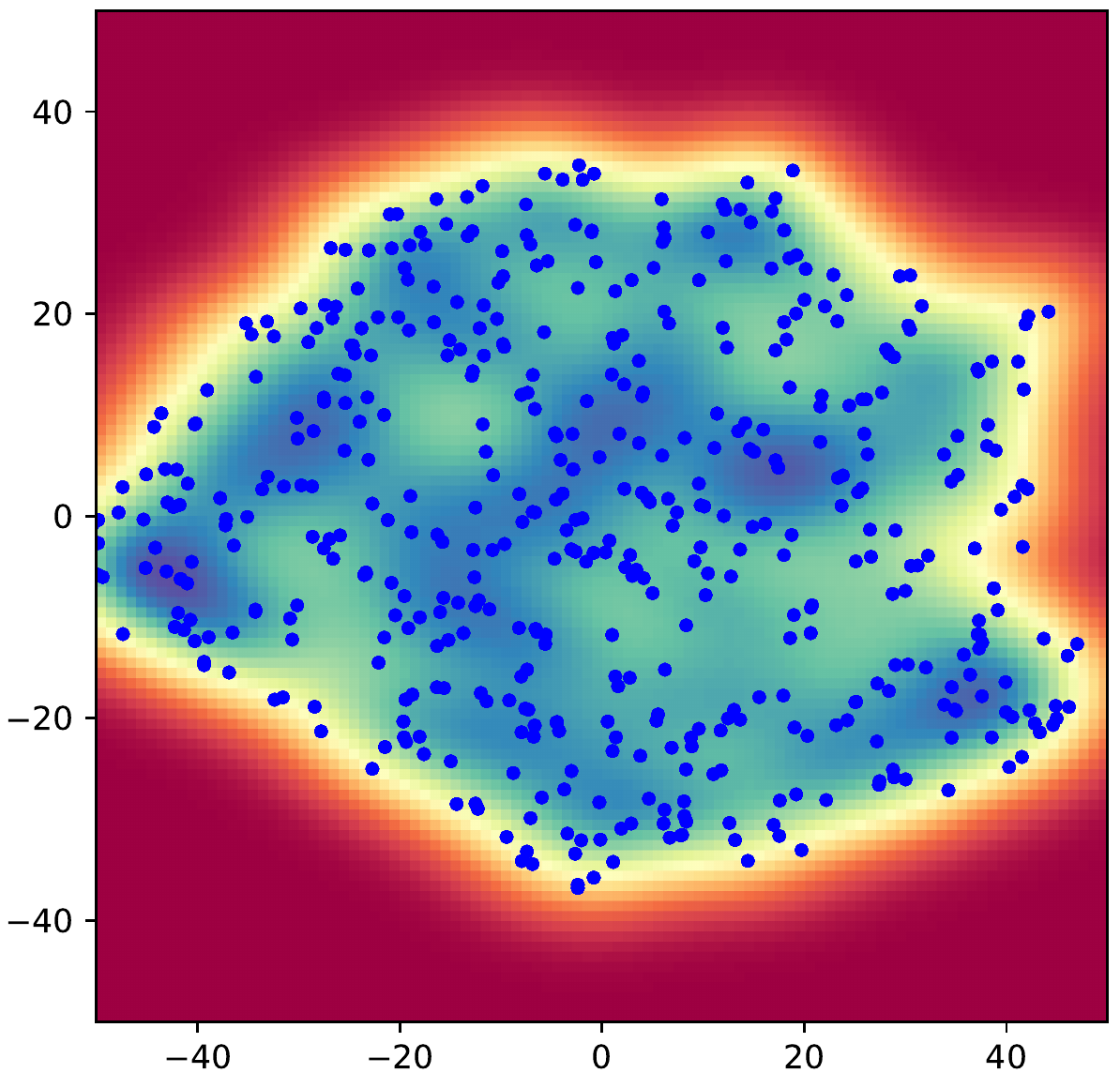}
    \caption{No contamination}
    \label{fig:roc_ucsd_ped1_frame}  
    \end{subfigure}
    \begin{subfigure}{0.19\textwidth}
        \includegraphics[width=\textwidth]{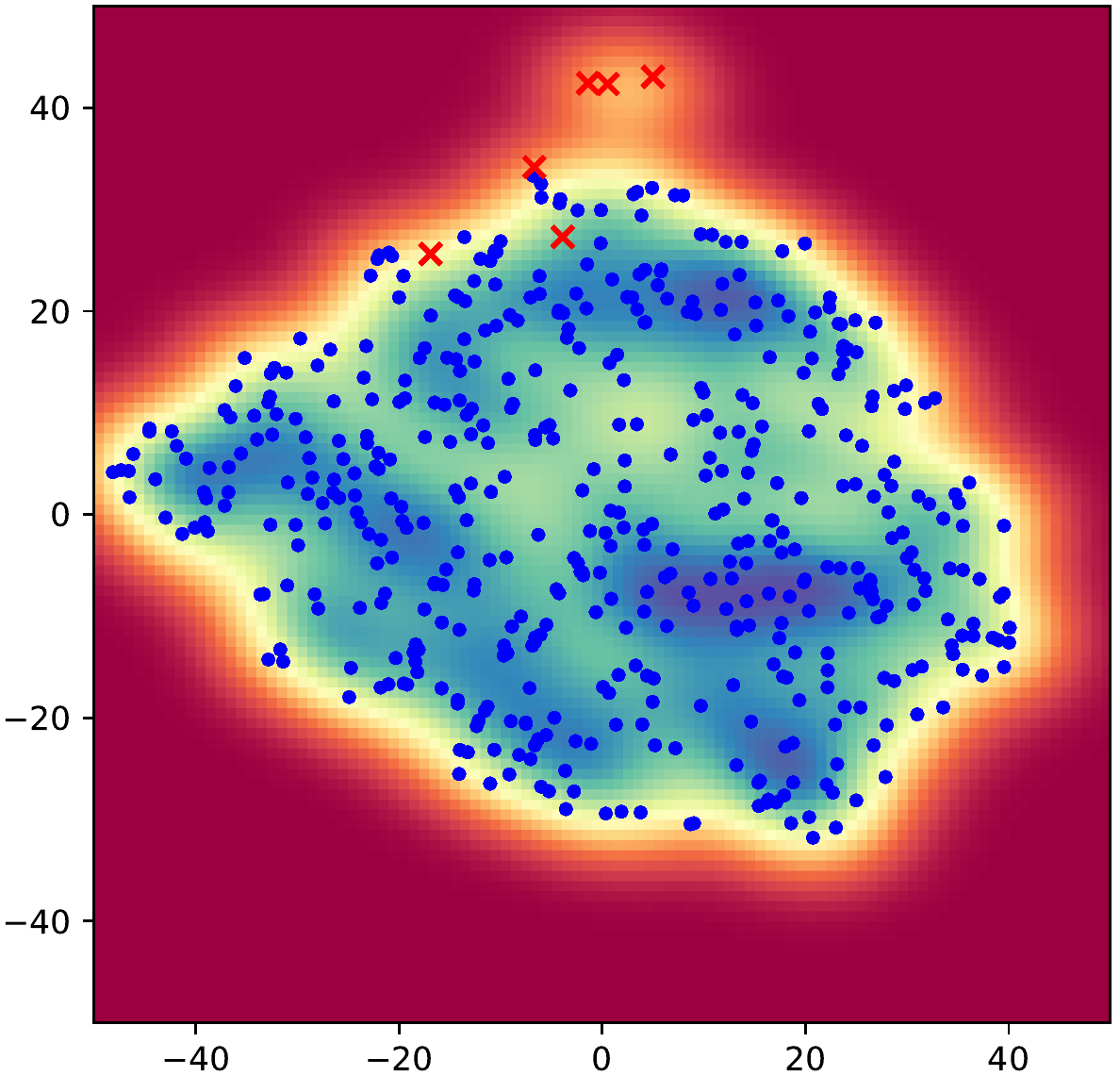}
    \caption{1\%}
    \label{fig:roc_ucsd_ped2_frame}  
    \end{subfigure}
    \begin{subfigure}{0.19\textwidth}
        \includegraphics[width=\textwidth]{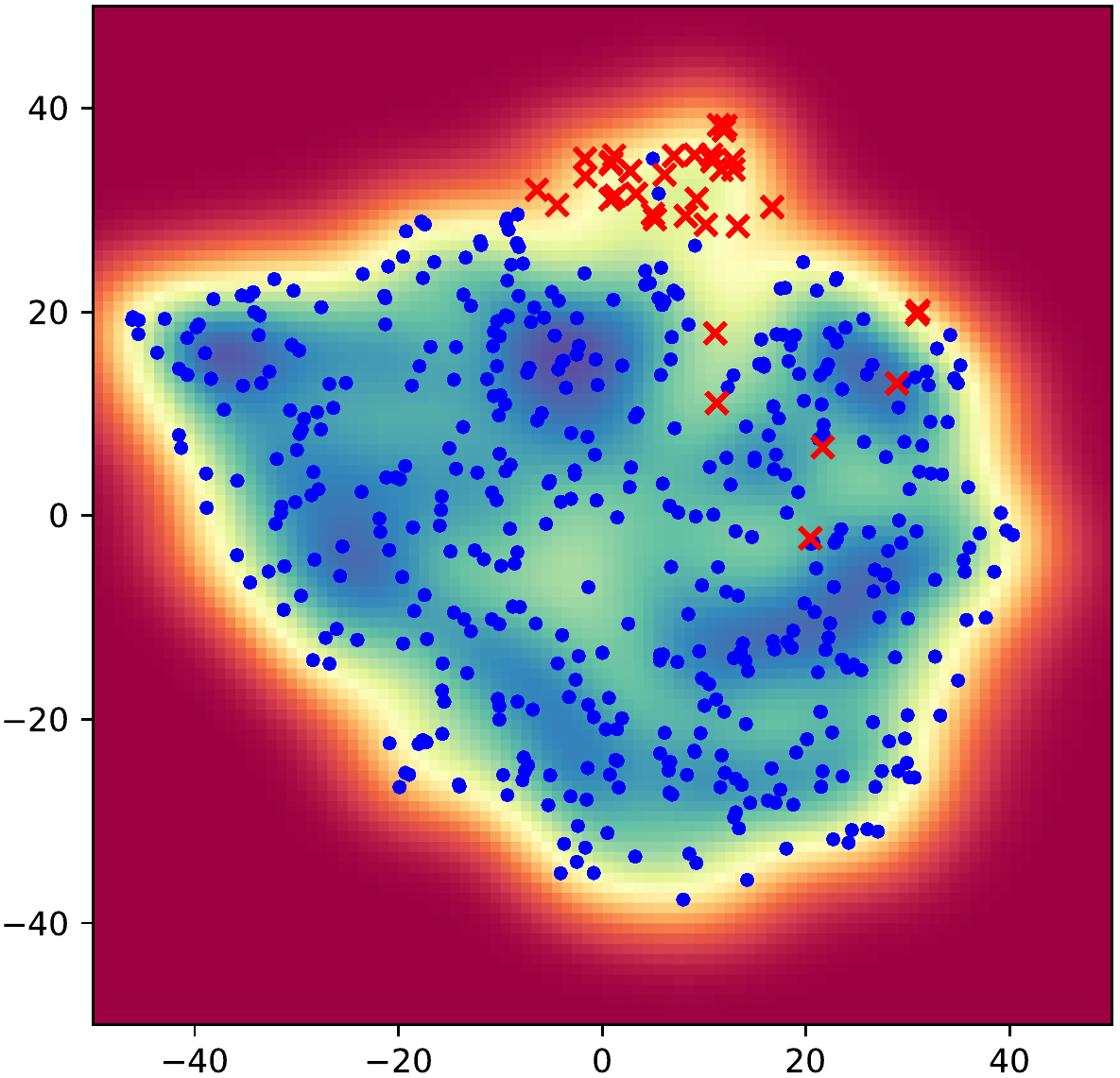}
    \caption{5\%}
         \label{fig:roc_ucsd_ped3_frame}  
    \end{subfigure}
    \begin{subfigure}{0.19\textwidth}
        \includegraphics[width=\textwidth]{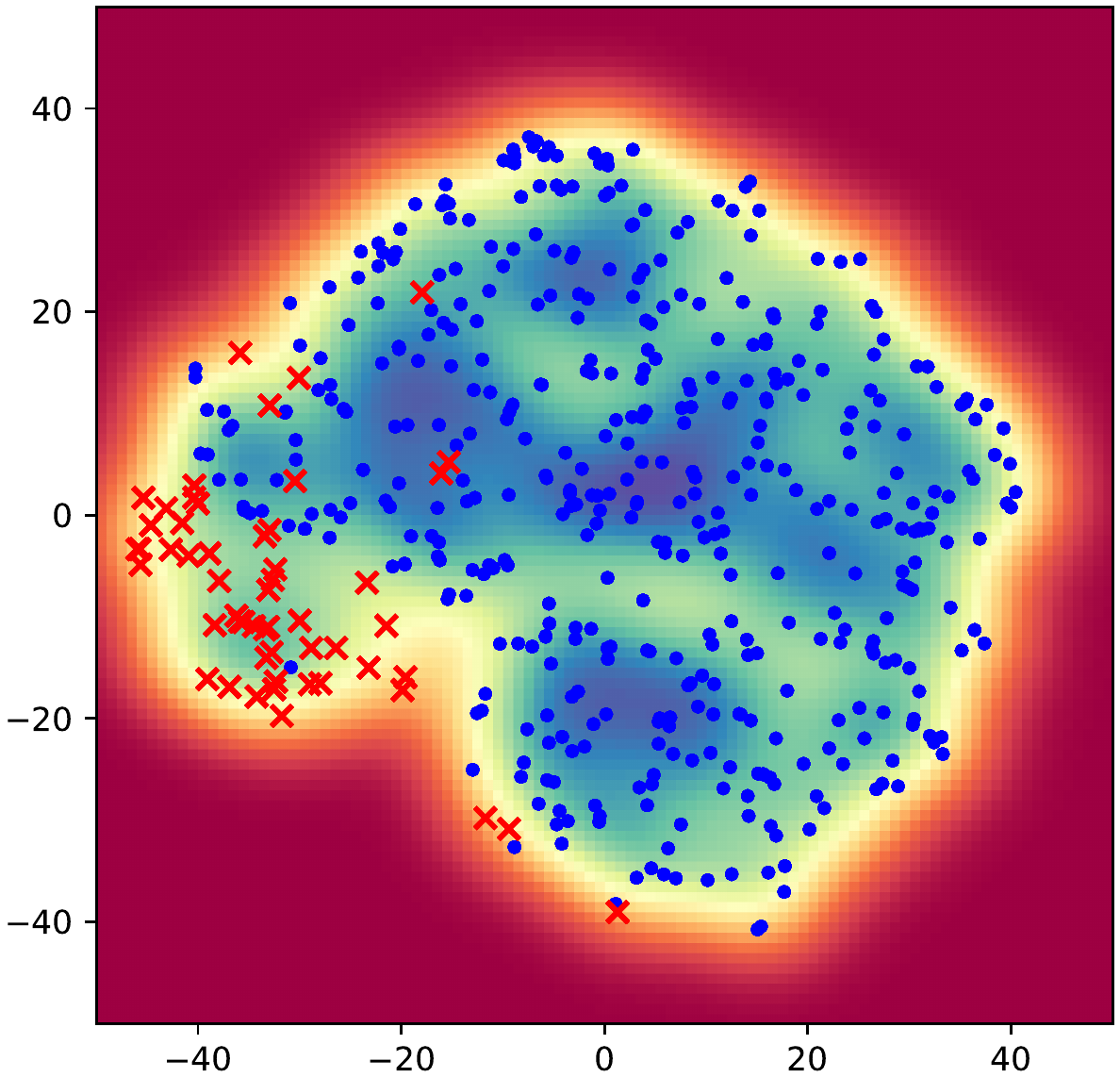}
    \caption{10\%}
         \label{fig:roc_ucsd_ped4_frame}  
    \end{subfigure}
    \begin{subfigure}{0.19\textwidth}
        \includegraphics[width=\textwidth]{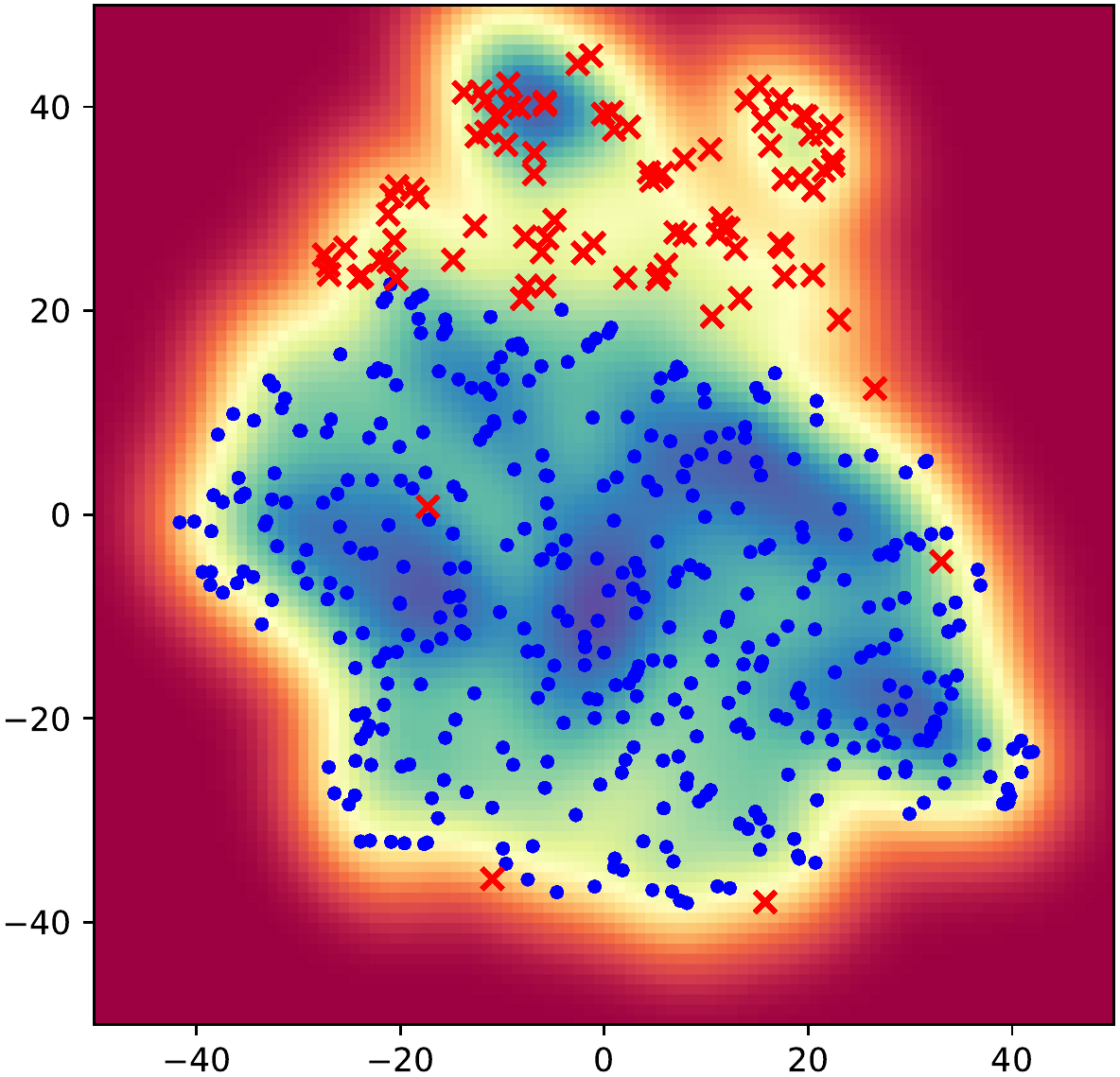}
    \caption{20\%}
         \label{fig:roc_ucsd_ped5_frame}  
    \end{subfigure}
    \caption{Entropy and distribution of latent features under different contamination ratios: (a) No contamination, (b) 1\%, (c) 5\%, (d) 10\%, and (e) 20\%. The samples on the `5' class on the MNIST dataset are used as normal (blue dots) and contaminated samples (red x-marks) are randomly picked from the training samples of the remaining classes. The 500 samples are randomly picked for the visualisation. When a dataset is highly contaminated (\ie contamination ratio over 10\%), contaminated samples are also located in a low entropy region.}
\label{fig:1}   
\vspace{-2ex}
\end{figure}

Various methods have been proposed \citep{ruff2019deep,song2017,akcay2018,chalapathy2019deep,zong2018deep} to improve the robustness of AD methods on contaminated datasets. Particularly, filtering contaminated samples based on contamination ratio \citep{zong2018deep,ruff2019deep}, semi-supervised learning approaches that uses explicit abnormal samples in the training step \citep{wang2005,liu2006,gornitz2013,ruff2019deep}, and contamination sample prediction approaches based on geometric distance measurement \citep{berg2019unsupervised,li2021deep,lai2020robust}, have been proposed. However, the aforementioned approaches are domain or data-type specific. 
Additionally, those methods assume that abnormal samples are likely to be located far from the distribution of normal samples, and the entropy of abnormal samples is higher than that of normal samples \citep{berg2019unsupervised,li2021deep,lai2020robust}.
Unfortunately, as shown in Figure \ref{fig:1}, if a training dataset is highly contaminated, the contaminated samples can also form a low entropy space by themselves.

In this paper, we present Normality-Calibrated Autoencoder (NCAE), which is robust to the training dataset contamination. Our key idea on the NCAE is to adversarially generate high confident normal samples from a low entropy feature space and then to contrastively compare the generated samples with the input samples for estimating contamination score. 
After identifying the contaminated samples, NCAE is trained to maximise reconstruction error of the found sample. 

\section{Normality-Calibrated Autoencoder}
\subsection{Learning normality-calibrated autoencoder}
For $n$ number of input samples with $D$ dimensions $\mathcal{X}=\{x_{i}\}_{i=1:n}$, $x\in\mathbb{R}^{D}$ and the corresponding latent features with $d$ dimensions $\mathcal{Z}=\{z_{i}\}_{i=1:n}$, $z\in\mathbb{R}^{d}$, let an autoencoder is composed of an encoder $f(x):x\xrightarrow{}z$ and a decoder $g(x):z\xrightarrow{}\bar{x}$. The general objective of the autoencoder is training $f$ and $g$ to minimise an error between input samples $x$ and the reconstruction results $\bar{x}$, as follows:
\begin{equation}
\begin{split}
\min_{f,g}\mathbb{E}_{x\sim{}p_{\mathcal{X}}}||x-\bar{x}||^{2},{\quad}\bar{x} = g\cdot{}f(x),
\end{split}
\label{eq:rec_loss}
\end{equation}
where $p_{\mathcal{X}}$ denotes the entire input samples. However, an autoencoder is known to have an over-confidence issue, \ie low reconstruction error of unseen samples. AD methods using the autoencoder usually identify abnormal samples using the reconstruction error. Therefore, even if the autoencoder takes anomaly samples as inputs, it may not distinguish whether the samples are abnormal or not \citep{pidhorskyi2018generative,yu2021abnormal}. This over-confidence issue would be more deepened when a training dataset is contaminated. 

One straightforward approach to prevent this issue is adding an extra term to maximise reconstruction error for contaminated samples. We define normality-calibrated reconstruction (NCR) loss as follows:
\begin{equation}
\begin{split}
\min_{f,g}\mathbb{E}_{x\sim{}p_{\mathcal{X}^{\text{N}}}}||x-\bar{x}||^{2}-\mathbb{E}_{x^{\text{c}}\sim{}p_{\mathcal{X}^{\text{C}}}}||x^{\text{c}}-\bar{x}^{\text{c}}||^{2},
\end{split}
\label{eq:ccr_loss}
\end{equation}
where $p_{\mathcal{X}^{\text{N}}}$ and $p_{\mathcal{X}^{\text{C}}}$ denote the normal samples and contaminated samples, respectively, among input samples $p_{\mathcal{X}}$. Now, we should find out which samples are contaminated to optimise autoencoder using \eqref{eq:ccr_loss} properly. 

\subsection{high-confidence normal samples generation using Generative Adversarial Network}
We find contaminated samples by using high confident normal samples generated from low entropy latent space. We apply the generative adversarial network (GAN) \cite{goodfellow2014} framework to do this. The high-confidence normal sample generation via the GAN framework is carried out as follows. 
Initially, we transform a distribution of all latent features, which are encoded from input samples through the encoder $f$, to a more knowledgeable probabilistic distribution such as Gaussian distribution. And then, we generate samples using noise signals sampled from the centre of the knowledgeable distribution, \ie the low entropy space.
An adversarial loss for transforming a latent feature distribution to a more knowledgeable probabilistic distribution is defined by follow: 
\begin{equation}
\begin{split}
\min_{f} \max_{D_{l}} \mathbb{E}_{\omega\sim{}N(\mu_{\mathcal{Z}},I_{d})}[\log{}D_{l}(\omega)] + \mathbb{E}_{x\sim{}P_{\mathcal{X}}}[\log{}(1-D_{l}(f(x)))],
\end{split}
\label{eq:adv_l}
\end{equation}
where $D_{l}$ denotes the discriminator for latent features, and $N(\mu_{\mathcal{Z}},I_{d})$ defines a Gaussian distribution with the mean of latent features $\mu_{\mathcal{Z}}\in\mathbb{R}^{d}$ and a covariance matrix defined by an identity matrix $I_{d}\in\mathbb{R}^{d\times{}d}$. $\mu_{\mathcal{Z}}$ is initialised by the mean value of latent features: $\mu_{\mathcal{Z}} = \frac{1}{n}\sum_{i=1}^{n}z_{i}$. We would want each component of $z$ to be maximally informative such as each of them to be an independent random variable. Therefore, the covariance matrix is determined by the $d\times{}d$ identity matrix.

Since as $f$ and $D_{l}$ are being updated, $\mu_{\mathcal{Z}}$ would be shifted during the training step. $\mu_{\mathcal{Z}}$ is updated at every training step as follows:
\begin{equation}
\begin{split}
\mu_{\mathcal{Z}}^{t+1} = \mu_{\mathcal{Z}}^{t}-\gamma\frac{1}{m}\sum_{i=1}^{m}(\mu_{\mathcal{Z}}^{t}-z_{i}), \quad \mu_{\mathcal{Z}}^{0} = \frac{1}{n}\sum_{i=1}^{n}z^{0}_{i}
\end{split}
\label{eq:mu}
\end{equation}
where $\mu_{\mathcal{Z}}^{t+1}$ and $\mu_{\mathcal{Z}}^{t}$ denote the $\mu_{\mathcal{Z}}$ on $t+1$-th and $t$-th training step, respectively. $m$ is the batch size and $z_{i}$ is $i$-th latent features on the batch. $\gamma$ is a learning rate. 

To generate high confident normal samples, we formulate the following adversarial loss:
\begin{equation}
\begin{split}
\min_{g} \max_{D_{s}} \mathbb{E}_{x\sim{}P_{\mathcal{X}}}[\log{}D_{s}(x)] + \mathbb{E}_{\acute{\omega}\sim{}N(\mu_{\mathcal{Z},\sigma{}I_{d}})}[\log(1-D_{s}(g(\acute{\omega}))],
\end{split}
\label{eq:adv_sample}
\end{equation}
where $D_{s}$ denotes the discriminator for samples, and $N(\mu_{\mathcal{Z}},\sigma I_{d})$ is a $d$-dimensional Gaussian distribution with the mean $\mu_{\mathcal{Z}}\in\mathbb{R}^{d}$ and the covariance matrix $\sigma I_{d}\in\mathbb{R}^{d\times{}d}$. $\mu_{\mathcal{Z}}$ is equivalent to the $\mu_{\mathcal{Z}}$ in \eqref{eq:adv_l}. $\sigma{}I_{d}$ is defined by multiplication of a scalar value $\sigma\in[0,1]$ and the identity matrix $I_{d}$. $\sigma$ is a hyperparameter to control the compactness of random noise for generating samples using the decoder $g$. The smaller $\sigma$ can give more chances to generate high confident normal samples by generating a feature close to the centre of the probability distribution.

\subsection{Contaminated sample mining and joint learning}
To predict contaminated samples, we use the generated high confident normal samples as a dictionary. With the generation process for high confident normal samples: $g(\acute{\omega})=\acute{x}$, we construct a latent feature dictionary $\mathcal{M} = [\acute{z}_{i}]_{i=1:m}$, $\acute{z}_{i} = f(\acute{x}_{i})$ and $\mathcal{M}\in\mathbb{R}^{m\times{}d}$, where $m$ is the batch size. By leveraging $\mathcal{M}$ and given each training batch $\{x_{i}\}_{i=1:m}$, we define a pseudo contamination score $c_{i}$ of each input sample $x_{i}$ as follow:
\begin{equation}
\begin{split}
   c_{i} = \frac{1}{m}\sum_{j=1}^{m}f(x_{i})\cdot\acute{z}_{j}^{\mathrm{T}},\quad \acute{z}_{j} \in \mathcal{M},
\end{split}
\label{eq:score}
\end{equation}
where $\mathrm{T}$ denotes the transpose of the vector. We apply $l2$-normalisation to improve robustness on the variation of the vector scale of the operation.

We predict the contaminated samples by sorting the score in descending order and picking top-$\tau$\% samples among the sorted results as the contaminated samples; thus, the number of predicted contaminated samples are decided by $\tau{}m$ that is a multiplication of $\tau$ and the batch size $m$. The above process is represented as follow:
\begin{equation}
\begin{aligned}
\nonumber
\begin{split}
    &\mathcal{X}^{\text{C}} = \{x_{t}\}_{t\in{}C[1:\lceil{}\tau{}m\rceil]},\quad C = \mathop{\arg \operatorname {sort}}_{i} c_{i},\quad w.r.t., 1 \le i \le m\\
\end{split}
\label{eq:prediction}
\end{aligned}
\end{equation}
where $C$ is a set of the sorted indices of input batch samples in descending order of the contamination score (\eqref{eq:score}), and $\mathcal{X}^{\text{C}}$ is a set of predicted contaminated samples. $\lceil\cdot\rceil$ denotes the ceiling function. $\tau$ effects of deciding the number of predicted contaminated samples, so it directly affects the AD performance of our method. 

The objective function for joint learning the entire components on our method is as follows:
\begin{equation}
\begin{aligned}
\begin{split}
\min_{f,g} \max_{D_{l},D_{s}} \;\;\; &\underbrace{\mathbb{E}_{x\sim{}p_{\mathcal{X}^{\text{N}}}}||x-f\cdot{}g(x)||^{2}-\mathbb{E}_{x\sim{}p_{\mathcal{X}^{\text{C}}}}||x-\bar{x}'||^{2}}_{\text{(a)}} \\
&+\underbrace{\mathbb{E}_{\omega\sim{}N(\mu_{\mathcal{Z}},I_{d})}[\log{}D_{l}(\omega)] + \mathbb{E}_{x\sim{}P_{\mathcal{X}}}[\log{}(1-D_{l}(f(x)))]}_{\text{(b)}} \\
&+\underbrace{\mathbb{E}_{x\sim{}P_{\mathcal{X}}}[\log{}D_{x}(\omega)] + \mathbb{E}_{\omega'\sim{}N(\mu_{\mathcal{Z},\sigma{}I_{d}})}[\log(1-D_{s}(g(\omega'))]}_{\text{(c)}},
\end{split}
\label{eq:joint}
\end{aligned}
\end{equation}
where $\bar{x}'$ is defined by the nearest sample from the given contaminated samples among the generated high confident normal samples $g(N(\mu_{\mathcal{Z},\sigma{}I_{d}}))$ on the latent feature space. (a), (b), and (c) denote the NCR loss and the two adversarial losses, respectively.

\section{Experiments}
\subsection{Experiment setting and Dataset}
We follow the unsupervised AD protocol described by Ruff \etal \citep{ruff2019deep}. MNIST and Fashion-MNIST datasets are used for the experiments. We set one of the classes provided by a dataset as normal and others as abnormal. After we decide contamination ration $\rho = \frac{A}{N+A}$, where $N$ and $A$ are the numbers of normal and abnormal samples, respectively, we pick normal samples from the chosen class and contaminated samples from the remaining classes. In the test phase, the samples of the normal class are labelled by 0, and other samples are labelled by 1. For the performance analysis, Receiver Operating Characteristic (ROC) curve and Area Under the Curve (AUC) are used. 

We employ LeNet-type convolutional neural networks (CNNs) on MNIST and Fashion-MNIST datasets, where each convolutional module consists of a convolutional layer followed by leaky ReLU activation functions with leakiness of 0.1. We use the Adam optimiser with the recommended default hyperparameters \citep{kingma2014}. The batch size is set to 128. The initial learning rate is 0.01 and decayed every 10 epochs by multiplying 0.1. $\sigma$ and $\tau$ are decided by 0.1 and 0.1 (based on the results from the ablation study), respectively.

\subsection{Ablation study}
We analyse unsupervised AD performance depending on the setting of $\sigma$ and $\tau$. MNIST and Fashion-MNIST datasets are used for the ablation study. Ablation studies are conducted based on the experimental protocol described in the previous section. The contamination ratio $\rho$ is fixed to 0.2.

\begin{figure}[t]
\centering
    \begin{subfigure}{0.48\textwidth}
        \includegraphics[height=3.9cm,width=\textwidth]{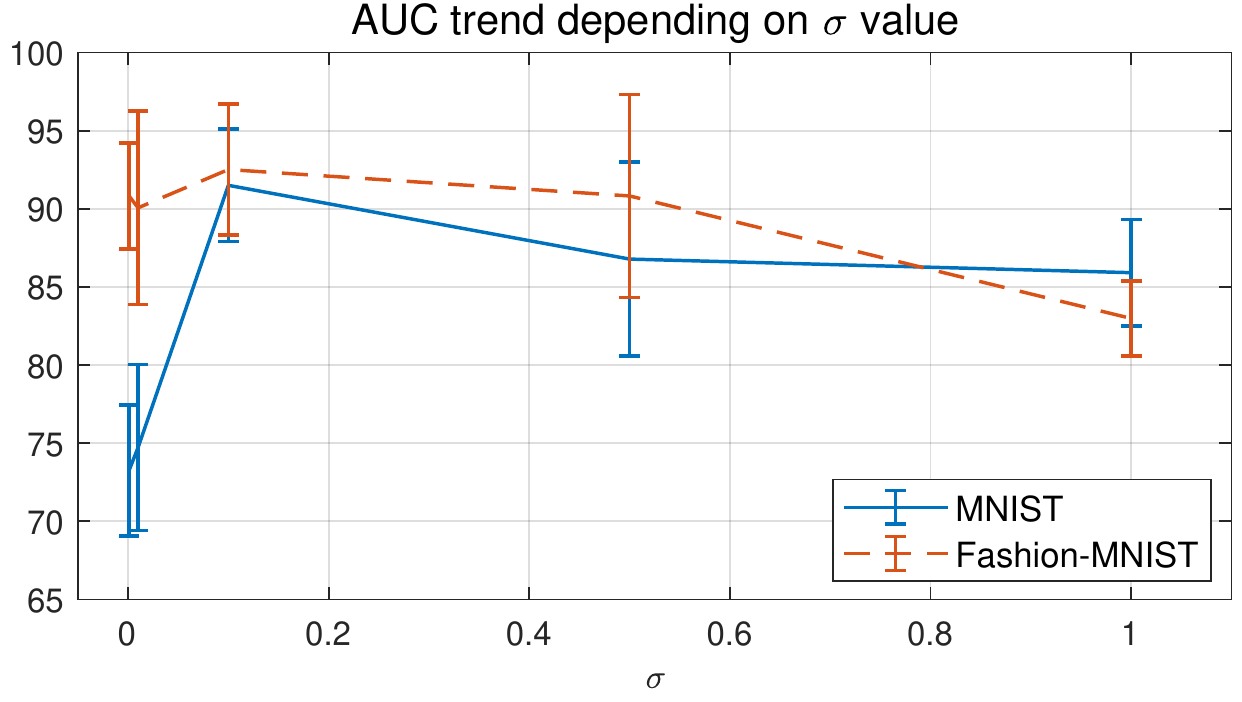}
        \caption{Effects of $\sigma$}
    \end{subfigure}
    \hspace{2ex}
    \begin{subfigure}{0.48\textwidth}
        \includegraphics[width=\textwidth]{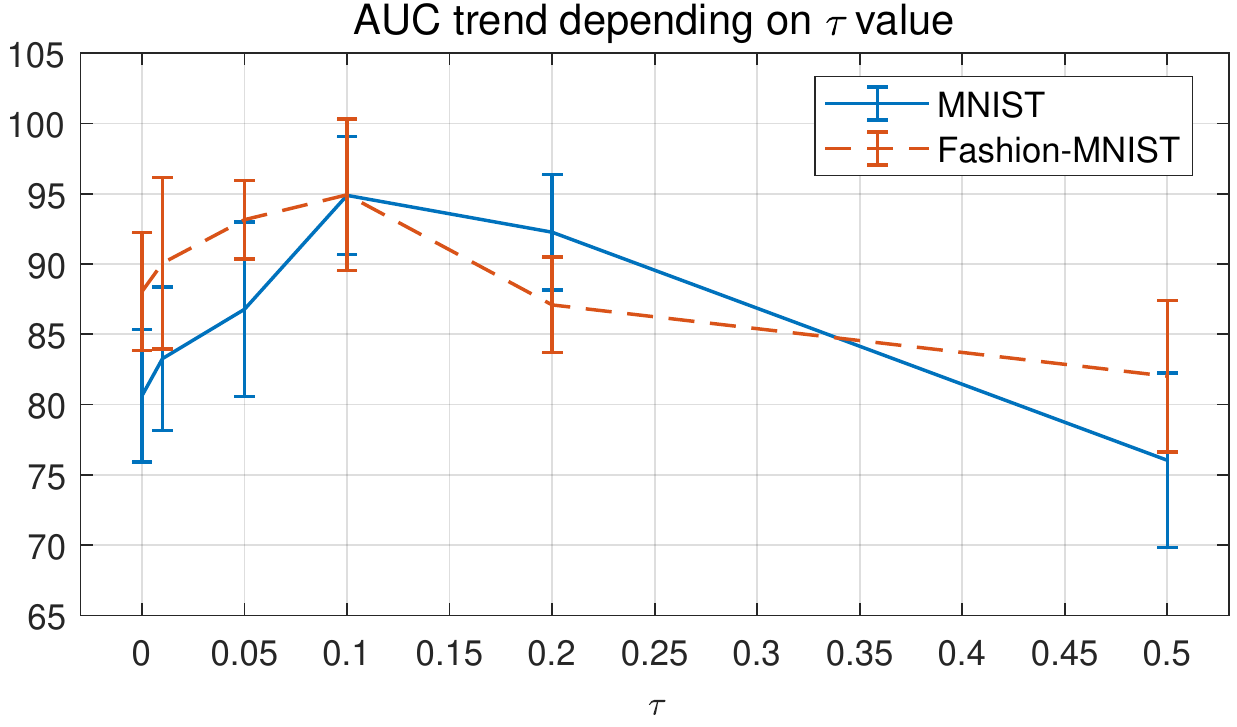}
        \caption{Effects of $\tau$}
    \end{subfigure}
    \caption{Ablation studies about unsupervised AD performance depending on $\sigma$ and $\tau$. (a) and (b) represent the trends of AUC with respect to the setting of $\sigma$ and $\tau$, respectively, on the MNIST and Fashion-MNIST datasets.}
\label{fig:2}   
\vspace{-2ex} 
\end{figure}

\textbf{Parameter analysis on $\sigma$:} When $\sigma$ is too small, then the distribution of sample noise for generating samples would be too compact so that the generated samples can not provide comprehensive information to cover the diverse patterns of normal samples. On the other hand, when $\sigma$ is too large, then there is a possibility that the noise can be sampled from low entropy space (\ie abnormal samples also can be generated). 

Figure \ref{fig:2}(a) shows the AUC trends depending on the $\sigma$. The AUC increases rapidly in the case of sigma is less than 0.1, and then decreases gradually. This can be interpreted as follows. If the sampling space is too compact (\ie when $\sigma$ is too small), it means that the generated normal sample does not provide enough information to distinguish the contaminated sample. When sampling space is too broad (\ie when $\sigma$ is too large), it also degrades performance, but 
the impacts of the braoder sampling space are relatively less than that of the smaller sampling space (e.g., when $\sigma \le 0.1$).
The best performance is obtained by $\sigma$ of 0.1.

\textbf{Parameter analysis on $\tau$:} $\tau$ decides the number of predicted contaminated samples per training batch. The lower $\tau$ can provide more precise prediction performance but may not enough to provide more comprehensive prediction performance. In contrast, when $\tau$ is too large, the predicted results possibly more accurate but also may have a lot of false-positive results.

As shown in Figure \ref{fig:2}(b), the AUC increases rapidly with $\tau$ from 0 to 0.1 and then decreases slowly. The results can be interpreted as follows. Finding contaminated samples themselves has a large impact on the AD performance, but the quantity of found samples affects less to AD performance. But, predicting too many samples may degrade AD performance by taking a great number of false positives. The best performance is obtained by $\tau$ of 0.1. 

\subsection{Comparison with other methods}
\label{sec:as}
We consider the OC-SVM \citep{scholkopf2001} 
, isolation forest (IF) \citep{liu2008}, and KDE \citep{parzen1962} for shallow unsupervised baselines. For deep unsupervised competitors, we consider general binary classifider (supervised), convolutional autoencoders (CAE), deep support vector data description (Deep SVDD) \citep{ruff2018}, semi-supervised anomaly detection (SSAD) \citep{ruff2019deep}, semi-supervised deep generative model (SS-DGM) \citep{kingma2014}, and deep semi-supervised anomaly detection (Deep SAD) \citep{ruff2019deep}. We repeat this training set generation process 10 times per AD setup over all the nine respective anomaly classes and report the average results over the resulting 90 experiments per contamination ratio.

Table \ref{tab:pollution} shows the quantitative performance comparison depending on the contamination ratio $\rho$. In the comparison using the MNIST dataset, the proposed NCAE achieves the best performances except when the dataset is not contaminated ($\rho=0.0$). Even compared with semi-supervised approaches \citep{ruff2018,ruff2019deep} which use explicit anomaly samples in the training phase, the NCAE shows outstanding performances. This trend is also shown in the performance comparison using the Fashion-MNIST dataset. The NCAE produces the AUC of 91.57 and 88.97 for the Fashion-MNIST dataset with 0.1 and 0.2 contamination ratios, respectively. Those figures are the best performance among the listed methods when a dataset is contaminated.

The interpretation of the relatively low performance on the uncontaminated dataset ($\rho=0.0$) is as follows. Basically, our method is derived under the assumption that a training dataset is contaminated. Therefore, even if the dataset is not contaminated, the NCAE tries to find some anomaly samples and maximise the reconstruction errors of the samples during the model training. This process degrades the performance of our methods as shown in the experimental results. This is a critical defect of our method;

Overall, the comparison results demonstrate the advantage of the proposed NCAE that can detect anomaly samples on data contamination without prior knowledge or explicit abnormal samples in the training phase.

\begin{table*}[t] 
        \centering 
        \captionsetup{type=table}
        \captionof{table}{Performance comparison on unsupervised anomaly detection in terms of various contamination ratios $\rho$. MNIST and Fashion-MNIST datasets are used for the comparison. The \textbf{bolded} figures indicate the best performances.}%
        \label{tab:pollution}
        \resizebox{\textwidth}{!}{
        \begin{tabular}{llcccccccccc}
        \toprule
                      \multirow{2}{*}{Dataset}        &    \multirow{2}{*}{$\rho$}      &   \multirow{2}{*}{OC-SVM}     &   \multirow{2}{*}{IF}    &  \multirow{2}{*}{KDE}  &  \multirow{2}{*}{CAE}   & \multirow{2}{*}{Deep SVDD}     & \multirow{2}{*}{SSAD}     &    \multirow{2}{*}{SS-DGM}         & \multirow{2}{*}{Deep SAD}      & \multirow{2}{*}{Classification} & \multirow{2}{*}{NCAE}   \\
                            
       	    &  &      &           &        &      &     &       &   &     & & \\
        \midrule
       \multirow{5}{*}{MNIST}			& .00 	& 96.0$\pm$2.9         & 85.4$\pm$8.7                  & 95.0$\pm$3.3             & 92.9$\pm$5.7      & 92.8$\pm$4.9      & \textbf{97.9$\pm$1.8}        & 92.2$\pm$5.6      & 96.7$\pm$2.4      & 94.5$\pm$4.6 & 94.0$\pm$4.2\\ 
        
        	 				& .01 	& 94.3$\pm$3.9        & 85.2$\pm$8.8                  & 91.2$\pm$4.9             & 91.3$\pm$6.1      & 92.1$\pm$5.1      & 96.6$\pm$2.4       & 92.0$\pm$6.0      & 95.5$\pm$3.3      & 91.5$\pm$5.9 & \textbf{97.2$\pm$5.2}\\ 
        	 				
        					& .05 	& 91.4$\pm$5.2          & 83.9$\pm$9.2               & 85.5$\pm$7.1             & 87.2$\pm$7.1      & 89.4$\pm$5.8      & 93.4$\pm$3.4     & 91.0$\pm$6.9      & 93.5$\pm$4.1      & 86.7$\pm$7.4  & \textbf{97.0$\pm$7.1} \\ 
        					
        	 				& .10 	& 88.8$\pm$6.0         & 82.3$\pm$9.5                & 82.1$\pm$8.5             & 83.7$\pm$8.4      & 86.5$\pm$6.8      & 90.7$\pm$4.4     & 89.7$\pm$7.5      & 91.2$\pm$4.9      & 83.6$\pm$8.2 & \textbf{92.6$\pm$5.7} \\ 
        	 				
        	 				& .20 	& 84.1$\pm$7.6        & 78.7$\pm$10.5               & 77.4$\pm$10.9            & 78.6$\pm$10.3     & 81.5$\pm$8.4      & 87.4$\pm$5.6     & 87.4$\pm$8.6      & 86.6$\pm$6.6      & 79.7$\pm$9.4 & \textbf{89.8$\pm$7.4}\\ 
        	 				
        \midrule
         \multirow{5}{*}{F-MNIST}           & .00	& 92.8$\pm$4.7         & 91.6$\pm$5.5              & 92.0$\pm$4.9            & 90.2$\pm$5.8      & 89.2$\pm$6.2      & \textbf{94.0$\pm$4.4}        & 71.4$\pm$12.7     & 90.5$\pm$6.5      & 76.8$\pm$13.2  & 91.5$\pm$9.7\\
        
        	 				& .01 	& 91.7$\pm$5.0          & 91.5$\pm$5.5                & 89.4$\pm$6.3            & 87.1$\pm$7.3      & 86.3$\pm$6.3      & 92.2$\pm$4.9       & 71.2$\pm$14.3     & 87.2$\pm$7.1      & 67.3$\pm$8.1  & \textbf{94.5 $\pm$4.7}\\
        	 				
        					& .05 	& 90.7$\pm$5.5         & 90.9$\pm$5.9                & 85.2$\pm$9.1          & 81.6$\pm$9.6      & 80.6$\pm$7.1      & 88.3$\pm$6.2       & 71.9$\pm$14.3     & 81.5$\pm$8.5      & 59.8$\pm$4.6  & \textbf{92.4$\pm$ 8.2} \\
        					
        	 				& .10 	& 89.5$\pm$6.1        & 90.2$\pm$6.3               & 81.8$\pm$11.2          & 77.4$\pm$11.1     & 76.2$\pm$7.3      & 85.6$\pm$7.0         & 72.5$\pm$15.5     & 78.2$\pm$9.1      & 56.7$\pm$4.1   & \textbf{91.5$\pm$5.7} \\
        	 				
        	 				& .20 	& 86.3$\pm$7.7        & 88.4$\pm$7.6                 & 77.4$\pm$13.6             & 72.5$\pm$12.6     & 69.3$\pm$6.3      & 81.9$\pm$8.1  & 70.8$\pm$16.0     & 74.8$\pm$9.4      & 53.9$\pm$2.9 & \textbf{88.9$\pm$9.2}\\
        	\midrule
        \bottomrule
        \end{tabular}
        }
\end{table*}

\section{Conclusion}
In this work, we have proposed NCAE that is a generative method for fully unsupervised anomaly detection on contaminated data. The experimental results have suggested that the NCAE outperforms existing methods for fully unsupervised anomaly detection with a large margin, and they have also provided competitive performances compared with semi-supervised methods using explicit abnormal samples to train their AD model. 

\section*{Acknowledgement}
This work was supported by Institute of Information \& communications Technology Planning \& Evaluation (IITP) grant funded by the Korea government(Ministry of Science and ICT; MSIT) (No. 2020-0-00833, A study on 5G based Intelligent IoT Trust Enabler).

\bibliographystyle{abbrvnat}
\bibliography{ref}

\end{document}